\documentclass[10pt,twocolumn,letterpaper]{article}

\usepackage[pagenumbers]{cvpr} % To force page numbers, e.g. for an arXiv version

\usepackage[dvipsnames]{xcolor}

\usepackage{amsmath,amsfonts,bm}

\def\eqref#1{equation~\ref{#1}}
\def\1{\bm{1}}

\def\va{{\bm{a}}}
\def\vb{{\bm{b}}}

\def\vp{{\bm{p}}}

\def\vz{{\bm{z}}}

\def\mP{{\bm{P}}}

\def\mS{{\bm{S}}}

\DeclareMathAlphabet{\mathsfit}{\encodingdefault}{\sfdefault}{m}{sl}
\SetMathAlphabet{\mathsfit}{bold}{\encodingdefault}{\sfdefault}{bx}{n}

\def\sD{{\mathbb{D}}}
\def\sR{{\mathbb{R}}}

\def\sX{{\mathbb{X}}}

\DeclareMathOperator*{\argmax}{arg\,max}
\DeclareMathOperator*{\argmin}{arg\,min}

\definecolor{cvprblue}{rgb}{0.21,0.49,0.74}
\usepackage[pagebackref,breaklinks,colorlinks,allcolors=cvprblue]{hyperref}
\usepackage{multirow}
\usepackage{array}
\usepackage{adjustbox}
\usepackage{pifont}
\usepackage{xspace}
\usepackage{graphicx}
\usepackage[skip=2pt,font=small]{caption}
\usepackage{algorithm}
\usepackage{algpseudocode}
\makeatletter
\newcommand{\NoIndentBold}[1]{%
  \par\noindent\textbf{#1}% Ensure no indentation and bold text
  \@afterindentfalse\@afterheading% Prevent further indentation in this block
}

\usepackage[accsupp]{axessibility}  % Improves PDF readability for those with disabilities.

\newcommand{\ours}{OTA\xspace}

\usepackage{color, colortbl}
\definecolor{codegreen}{rgb}{0,0.74,0.38}
\definecolor{codegray}{rgb}{0.5,0.5,0.5}
\definecolor{codepurple}{rgb}{0.58,0,0.82}
\definecolor{backcolour}{RGB}{230, 230, 230}
\definecolor{LightCyan}{rgb}{0.88,1,1}
\definecolor{LightRed}{RGB}{255, 204, 203}
\newcommand{\transrowcolor}{\color{Gray!}}

\def\paperID{6633} % *** Enter the Paper ID here
\def\confName{CVPR}
\def\confYear{2024}

\title{Optimal Transport-Guided Source-Free Adaptation for Face Anti-Spoofing}

\author{
Zhuowei Li\textsuperscript{1}\thanks{Equal contribution.}\hspace{0.3em}\thanks{Work done during internship at AWS AI Labs.}\hspace{0.3em}, \ 
Tianchen Zhao\textsuperscript{2}\footnotemark[1]\hspace{0.3em}, \ 
Xiang Xu\textsuperscript{2}, \ 
Zheng Zhang\textsuperscript{2}, \ 
Zhihua Li\textsuperscript{2}, \\
Xuanbai Chen\textsuperscript{2}, \ 
Qin Zhang\textsuperscript{2}, \ 
Alessandro Bergamo\textsuperscript{2}, \ 
Anil K. Jain\textsuperscript{2}, \ 
Yifan Xing\textsuperscript{2} \\[0.5ex]
\textsuperscript{1}Rutgers University \quad 
\textsuperscript{2}AWS AI Labs \\[0.5ex]
}

\begin{document}
\maketitle
\begin{abstract}
Developing a face anti-spoofing model that meets the security requirements of clients worldwide is challenging due to the domain gap between training datasets and diverse end-user test data.
Moreover, for security and privacy reasons, it is undesirable for clients to share a large amount of their face data with service providers. In this work, we introduce a novel method in which the face anti-spoofing model can be adapted by the client itself to a target domain at test time using only a small sample of data while keeping model parameters and training data inaccessible to the client. Specifically, we develop a prototype-based base model and an optimal transport-guided adaptor that enables adaptation in either a lightweight training or training-free fashion, without updating base model's parameters.
Furthermore, we propose geodesic mixup, an optimal transport-based synthesis method that generates augmented training data along the geodesic path between source prototypes and target data distribution. This allows training a lightweight classifier to effectively adapt to target-specific characteristics while retaining essential knowledge learned from the source domain.
In cross-domain and cross-attack settings, compared with recent methods, our method achieves average relative improvements of 19.17\% in HTER and 8.58\% in AUC, respectively.
\end{abstract}    
\section{Introduction}
\label{sec:intro}

\begin{figure}[t]
  \centering
   \includegraphics[width=0.90\linewidth]{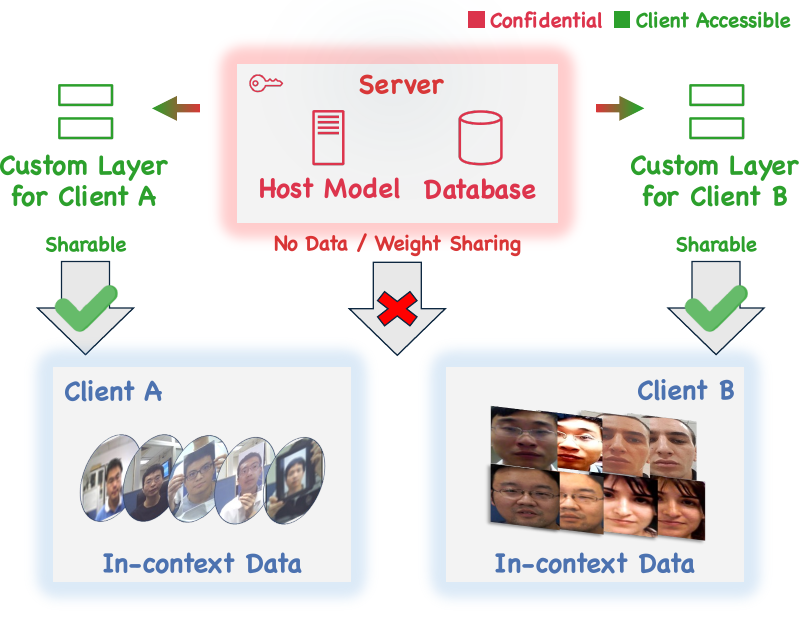}
   \caption{Overview of our setting. Unlike Few-shot Learning (FSL), Domain Adaptation (DA), and Domain Generalization (DG), our approach uses few-shot client data at test stage to adapt a customizable layer for each client’s needs, while keeping the host model and source training data confidential. This layer refers to prototypes and a lightweight classifier in our training-free and lightweight training approaches.}
   \vspace{-5mm}
   \label{fig:setting}
\end{figure}

Traditional approaches for developing face anti-spoofing models involve collecting extensive training datasets designed to cover a broad spectrum of real user interactions and spoof attempts. These datasets aim to represent diverse environmental conditions, user behaviors, demographic variations, and client-specific properties, such as artifacts introduced by image acquisition devices. However, these approaches have proven ineffective in practice~\cite{xu2024principles}, as the variety of user conditions is virtually limitless. Moreover, client-specific properties are dynamic and evolve over time, for example, upon the introduction of new image sensors and acquisition technologies. The conventional strategies struggle to keep up with these rapid advancements, as collecting new training datasets and retraining models on a daily basis is impractical.

Furthermore, conventional approaches involving tuning the model to meet client requirements using a small set of labeled end-user examples face several critical challenges. First, legal and privacy restrictions concerning human faces and security models often limit the quantity of available reference data and prohibit explicit sharing of host model parameters and source training data with clients for improvement. Second, fine-tuning a separate model at the service host side to address specific requirements for each client is impractical, as it is costly to maintain and manage future updates~\citep{deshpande2021linearized, koch2015siamese, snell2017prototypical}. Third, clients often desire a quick customization service that can avoid the inefficiencies of back-and-forth communication with the host in order to meet ever-evolving changes in their use cases.

In this work, we consider the problem of building a privileged system that allows for lightweight user-specific customization at testing time by the client, using only a few labeled samples, without the need for the host to share model parameters or training samples. These samples might be from different domains (cross-domain) or represent new spoofing instruments by adversaries (cross-attack). This differs from traditional Domain Adaptation (DA) and Domain Generalization (DG) methods, where DA requires both source and target domain data during training~\cite{li2018unsupervised, zhou2022generative, liu2022source, wang2021self} and DG relies solely on static parameters to generalize to downstream scenarios~\cite{jia2020single, chen2021generalizable, liu2023towards, zhou2024test, Zhou_2024_CVPR, huang2022adaptive, liu2024cfpl, sun2023rethinking}.

We address these challenges by introducing a customizable layer on top of the host model that can adapt to the target domain using a limited amount of client data, without accessing the source training data and host model parameters.
This customizable layer can be maintained and updated by the client in a source-free fashion, providing maximum flexibility to address each client's specific requirements~\cite{thapa2022splitfed}. The adaptation is realized by first learning, during the model training phase, a set of source prototypes that encode information about distributions of source domain data. At test time, we adapt the model through optimal transportation (OT) of the learned prototypes either in a training-free fashion or using a lightweight training method.
OT is particularly suitable for our application where the size of empirical data available to represent the true underlying distribution is limited~\citep{cuturi2013sinkhorn,ferradans2014regularized}, as it leverages Wasserstein distance to effectively exploit the geometry of the underlying feature space, faithfully aligning source and target distributions even when the data is sparse or unevenly sampled.

Concretely, in our training-free approach, we map source prototype features into the target domain with an optimal transport transformation that requires no learnable parameters. This transformation aligns the source prototypes with the target distribution's structure, allowing them to capture the unique characteristics of the target domain and make inference with the target data features.
Furthermore, inspired by our training-free approach, we propose an alternative method that trains a lightweight classifier on top of the frozen feature extractor, taking as input the source prototypes and a few target data features. 
To improve the learning of the decision boundary under a low-data regime, we introduce geodesic mixup, an OT-based synthesis method that generates augmented training data. Unlike traditional mixup data augmentations~\cite{zhang2017mixup,verma2019manifold} that perform point-wise linear interpolation between pairs of data features, the synthetic data generated along the geodesic path between source and target distributions guides the classifier to better capture the underlying feature manifold of both domains. By training on these samples, the classifier learns how features transition between domains, adapting to target-specific characteristics while maintaining knowledge about the source domain.

In summary, our main contribution is a novel framework for face anti-spoofing that uses a source-free few-shot adaptation approach based on prototypes and does not modify the parameters of the backbone host model. Our method leverages optimal transport for adaptation (OTA), where we employ the Wasserstein distance to estimate the true underlying data distribution based on a sparse sample of data, and a novel data augmentation method, geodesic mixup, to improve domain generalization performance. Our method achieves average relative improvements of 19.17\% in HTER and 8.56\% in AUC under cross-domain and cross-attack scenarios compared to state-of-the-art methods.

\section{Related Works}
\label{sec:related_work}

\noindent\textbf{Face anti-spoofing (FAS)} is a crucial security component in face recognition systems that has been extensively researched. Early efforts relied on handcrafted features like LBP~\cite{li2004live,de2013lbp,boulkenafet2015face}, HOG~\cite{yang2013face, komulainen2013context}, SIFT~\cite{patel2016secure}, and others~\cite{pan2007eyeblink,kollreider2007real,chetty2010biometric,xu2021improving}.
Deep learning has significantly advanced the field, with CNN-based methods~\cite{liu2019deep,liu2020disentangling,zhao2021learning,Wang_2022_CVPR,huang2022adaptive,liu2022spoof,xu2024principles} integrating feature extraction and classification into unified frameworks. However, many approaches struggle to generalize across domains in cross-dataset evaluations. Domain Adaptation (DA)~\cite{li2018unsupervised,xu2019d,wang2020unsupervised,guo2022multi,zhou2022generative,wang2019improving,liu2022source,liu2024source} adapts models from source to target domains using labeled or unlabeled target data~\cite{li2018unsupervised, zhou2022generative, liu2022source, wang2021self}, but it requires access to both source and target data at the training time, which is infeasible in our setting due to privacy concerns.
Domain Generalization (DG) leverages multiple source domains to train models that generalize to unseen domains~\cite{jia2020single, chen2021generalizable, liu2023towards, zhou2024test, Zhou_2024_CVPR, huang2022adaptive, liu2024cfpl, sun2023rethinking}, but it underperforms for significant domain shifts. 
Some methods~\cite{Rui2019Multi,jia2020single,Wangarticle} employ adversarial training for domain-invariant features, while others~\citep{shao2020regularized,chen2021generalizable,wang2021self,wang2022domain,sun2023rethinking} use meta-learning and domain separability to improve generalization.

Recent DG advances include instance-level domain-specific strategies like CIFAS~\cite{liu2022causal}, which uses causal intervention to mitigate domain bias, and AMEL~\cite{zhou2022adaptive}, which incorporates domain-specific features. Recently, \citet{liu2023towards} proposed unsupervised DG frameworks that leverage unlabeled data to learn generalizable features. While DA and DG methods have achieved success in FAS, both fall short in addressing practical scenarios where hosts must offer tailored solutions to clients with only a few labeled target samples at test time, without explicitly sharing the source model parameters. Our work investigates this constrained setting to highlight its practical importance and to draw research attention to this under-explored topic.\\

\noindent\textbf{Source-Free Domain Adaptation (SFDA)}~\citep{kim2021domain} decouples domain adaptation from direct source data usage by leveraging pre-trained source models. \citet{liang2020we} learns target-specific feature extraction with pseudo labels guided by class-wise prototypes. \citet{zhang2022few} and \citet{lee2023few} also consider few-shot SFDA scenarios, but they do not restrict the access to the source model. In FAS, \citet{lv2021combining} and \citet{mao2024weighted} employ pseudo labels for self-training but suffer from accumulated errors. \citet{liu2022source} considers a similar setting but also allows for unrestricted host model access. \citet{huang2023test} focuses on online adaptation to unlabeled target data without source data. Our method provides a lightweight solution for source-free few-shot adaptation without modifying the backbone host model parameters.\\

\begin{figure*}[t]
  \centering
   \includegraphics[width=\linewidth]{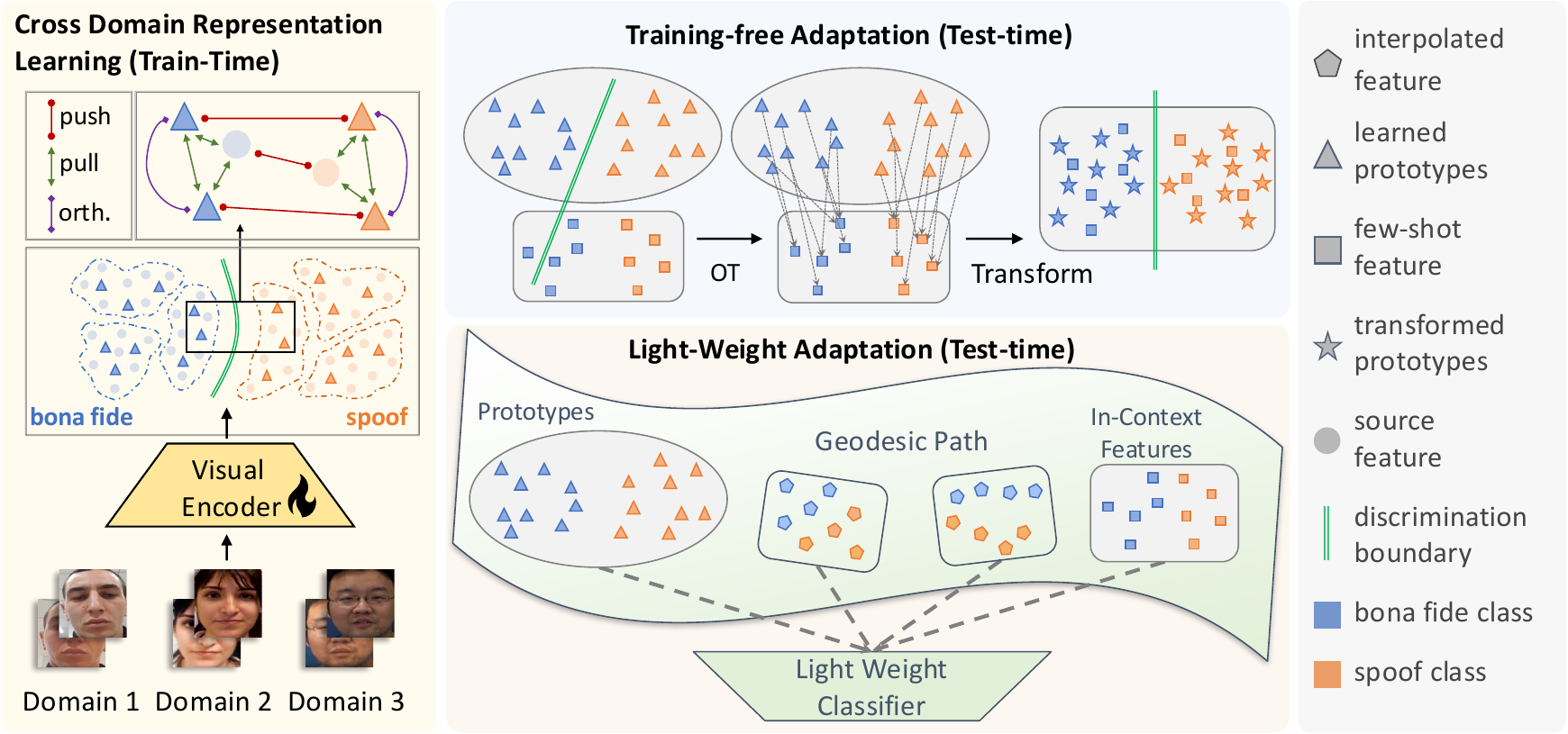}
   \caption{Overview of \ours. \ours learns a feature extractor and prototypes across multiple source domains during training. At test time, it adapts to few-shot client data via two approaches: a training-free method using optimal transport to shift source prototypes without learnable parameters, and a lightweight method training a classifier on synthetic data generated along the geodesic path, preserving source-domain understanding while adapting to target specifics.}
   \label{fig:overview}
   \vspace{-5mm}
\end{figure*}

\noindent\textbf{Optimal Transport (OT)} has been widely adopted in domain adaptation~\cite{courty2016optimal, zhao2019learning, le2019deep, lee2019sliced, shen2018wasserstein, balaji2019normalized, xu2019wasserstein, montesuma2021wasserstein} and generative modeling~\cite{arjovsky2017wasserstein,gulrajani2017improved} due to its ability to compute Wasserstein distances between probability distributions by effectively exploiting the underlying metric space geometry. OT is well-suited to our setting because its theoretical formulation adapts straightforwardly to discrete cases, working directly with empirical distribution estimates without assumptions on source and target distribution supports~\cite{courty2014domain,courty2016optimal,zhao2019learning,lee2019sliced}. Regularized OT addresses overfitting when few samples are available~\citep{cuturi2013sinkhorn,ferradans2014regularized}. While Mix-up~\cite{zhang2017mixup,verma2019manifold} has been used in Adversarial Domain Adaptation~\cite{xu2020adversarial,yan2020improve} to improve discriminator decision boundaries, our proposed geodesic mix-up serves a different purpose: it generates synthetic distributions by interpolating along a geodesic path in feature space with OT, and we train on data sampled from these generated synthetic distributions to learn feature transitions between domains, adapting to target characteristics while maintaining source domain understanding.
\citet{xu2022few} also considers applying OT to the FAS task; nevertheless, they employ linear mix-up to augment the training set before applying JDOT~\cite{courty2017joint}, a well-established DA method, which stands in contrast to our proposed geodesic mix-up.

\section{Methodology}
\label{sec:method}

We propose optimal transport-guided source-free few-shot adaptation for face anti-spoofing (\ours), which trains a privileged model that supports convenient customization at the test stage by either host or client. Different from standard classification training, \ours is built upon a prototype-based training framework~\citep{tanwisuth2021prototype} that learns a set of source prototype features as both last-layer classifiers and surrogates for source domain distributions. A detailed description of this framework is provided in Section~\ref{sec:pt_framework}. At the testing stage, \ours offers two approaches for test-stage adaptation through optimal transportation (OT) of the learned prototypes: a training-free method and a lightweight training method, introduced in Section~\ref{sec:ot_adapt} and in Section~\ref{sec:geodesic}, respectively. We refer to Fig.~\ref{fig:overview} for an overview of \ours.

\subsection{Problem Formulation}

During the training stage, we have access to $N$ labeled source-domain datasets: $\{\sD_i\}_{i=1}^{N}$, where each dataset $\sD_i$ consists of $M_i$ labeled training samples $\{(x_{ij},y_{ij})\}_{j=1}^{M_i}$. Here, $x_{ij}\in \sX_i$ is an image from the $i^\text{th}$ dataset and $y_{ij}\in\{0,1\}$ is its corresponding binary label, indicating whether the image is bona fide or a spoof, respectively. At test time, we assume the existence of a few-shot labeled dataset $\sD_t=\{(x_{tj},y_{tj})\}_{j=1}^{M_t}$ from the target domain, which is not accessible during the training of the feature extractor $f$. The size of $\sD_t$ is significantly smaller than that of each source dataset, \textit{i.e.}, $M_t \ll M_i, i\in\{1,2,...,N\}$. By default, we assume $\sD_t$ contains samples from both bona fide and spoof classes; we will investigate the one-class scenario in our ablation studies. The objective is to build a customizable layer on top of $f$ that adapts to the target domain using \textbf{$\sD_t$} during the adaptation phase of the test stage, without using $\{\sD_i\}_{i=1}^{N}$ or modifying $f$. 

\subsection{Prototype-based Framework}
\label{sec:pt_framework}
The primary challenge in our setting arises from the lack of explicit access to the source-domain data and host model at the test stage due to privacy and proprietary considerations. To this end, we adopt a prototype-based framework. Instead of learning a classifier, we train a multi-centroid prototype for each class, with $\vp^{\text{bona fide}} \in \sR^{D\times K}$ for the bona fide class and $\vp^{\text{spoof}} \in \sR^{D\times K}$ for the spoof class, where $K$ denotes the number of sub-centers (determined prior to training) and $D$ indicates the feature dimension.

Given an image embedding $\vz = f(x) \in \sR^D$, classification is performed by calculating the mean cosine similarity between $\vz$ and each set of prototypes. The label is then assigned based on the highest mean similarity over the $K$ prototypes:
\begin{equation}
c = \argmax_{c\in\{\text{bona fide}, \text{spoof}\}} \frac{1}{K}\sum_{k=1}^{K} \frac{\vz \cdot \vp_{k}^c}{\lVert \vz \rVert_2 \lVert \vp_{k}^c \rVert_2}.
\end{equation}
In this framework, prototypes not only serve as classifiers but also encapsulate the feature distributions of the source domains. We use prototypes with sub-centroids to improve their expressiveness. At the test stage, the feature extractor $f$ is treated as a black box that takes as input the test image and generates its corresponding feature vector.
The learned prototypes $\mP=\{\vp^{\text{bona fide}}, \vp^{\text{spoof}}\}\in\sR^{D\times K \times 2}$ are accessible since they are lightweight and respect privacy.

\noindent\textbf{Training.} Intuitively, an embedding feature $\vz_i$ of image $x_i$ should be close to its corresponding prototype sub-centroids while being distinctly separated from mismatched prototypes. Inspired by the ArcFace loss~\cite{deng2019arcface}, we enforce an additive margin $m$ in the angular space. Let $\mS \in \sR^{2\times K} = \mP^T\vz_i$ represent the centroid-wise similarity between an image embedding $\vz_i \in \sR^{D}$ and the prototypes $\mP \in \sR^{D\times K \times 2}$. We minimize the following loss function:
\begin{equation}
    \mathcal{L}_{\text{proto}} = - \frac{1}{|I|}\sum_{i \in I}\log \frac{e^{s \cos (\theta_{y_i} + m)}}{e^{s \cos (\theta_{y_i} + m)} + e^{s \cos \theta_{1-y_i}}},
    \label{eq:proto}
\end{equation}
where $\theta_{y_i} = \arccos \Bigl( \frac{1}{K} \sum_{k=1}^{K} \bigl( \mP_{y_ik}^T \vz_i \bigr) \Bigr)$, $y_i \in \{0,1\}$ is the label for the image $x_i$, and $I=\{1,2,\ldots,B\}$ is the index set of a batch. Unlike the original ArcFace loss, which minimizes the angular distance between a sample and its nearest sub-center, we reduce the cosine distance between a data embedding and its corresponding group of sub-centroids. The margin $m$ is set as a learnable parameter. To avoid trivial mode collapse among sub-centroids, we regularize sub-centroids via a class-wise orthogonal loss: $\mathcal{L}_{\text{orth}} = \bigl\| \vp^T \vp - \mathbf{I} \bigr\|_2^2$, where $\mathbf{I}$ is the identity matrix.

The loss function in Eq.~\ref{eq:proto} establishes the relationship between data instances and their corresponding prototypes. To further improve intra-class alignment and inter-class separability, as well as preserve the geometric information among different source domains, we introduce an instance-wise supervised contrastive loss~\cite{khosla2020supervised}:
\begin{equation*}
\mathcal{L}_{\text{con}} =
    \sum_{i \in I_{\text{multi}}} \frac{-1}{|P(i)|} \sum_{p \in P(i)} \log \frac{\exp(\vz_i \cdot \vz_p / \tau)}{\sum_{a \in A(i)} \exp(\vz_i \cdot \vz_a / \tau)},
\end{equation*}
where $I_{\text{multi}}=\{1,2,\ldots,2B\}$ is the index set of the multiviewed batch~\cite{khosla2020supervised}, $ A(i) = I / \{i\}$ and $P(i) = \{ p \in A(i):\tilde{y}_p = \tilde{y}_i \}$. Our use of supervised contrastive loss is two-fold: a coarse supervision $\mathcal{L}_{\text{con}}^{\text{coarse}}$ involving only two classes—bona fide and spoof—and a fine-grained version $\mathcal{L}_{\text{con}}^{\text{fine}}$ that treats different source domains and attack methods as distinct sub-classes. Our final learning objective is:
\begin{equation}
    \mathcal{L} =  \mathcal{L}_{\text{proto}} + \alpha \mathcal{L}_{\text{con}}^{\text{coarse}} + \beta \mathcal{L}_{\text{con}}^{\text{fine}} + \eta \mathcal{L}_{\text{orth}},
\end{equation}
where the scalars $\alpha$, $\beta$, and $\eta$ regulate the strength of the coarse contrastive loss, fine-grained contrastive loss, and orthogonal term, respectively.

\subsection{In-Context Adaptation Using Optimal Transportation}
\label{sec:ot_adapt}

\ours\ considers a scenario where only the prototypes $\vp$ and a few target-domain examples $\sD_t$ are available at the adaptation time during the test stage. 
Our approach is to identify a transformation function that can adjust the prototypes $\vp$ appropriately based on $\sD_t$. A critical property of prototypes is that they encode rich information about the source domains within their geometric structure. Motivated by this, we propose to use Optimal Transportation (OT), which respects the geometries of both source and target distributions~\cite{villani2009optimal}, to generate the transformation function.

Specifically, we derive this transformation function by solving the following regularized OT optimization problem:
\begin{equation}
\begin{split}
    \gamma^* = \argmin_{\gamma \in \mathbb{R}_+^{2K \times M_t}} \sum_{i,j} \gamma_{i,j} M_{i,j} + \lambda \Omega_{\alpha}(\gamma) \\ 
    \text{s.t.} \; \gamma \mathbf{1} = \va; \; \gamma^T \mathbf{1} = \vb; \; \gamma \geq 0,
\end{split}
\end{equation}
where $\gamma^*$ represents the optimal transportation plan, and $\va, \vb$ are the weights (summing to 1) of the source and target distributions, respectively. $M$ is the cost matrix of size $K \times M_t$, where $2K$ is the number of prototypes, and $M_t$ is the number of data points in the target domain. Each entry $M_{i,j}$ denotes transportation cost from source unit $i$ to target unit $j$. Here, unit $i$ refers to a sub-center from learned prototypes, and $j$ corresponds to a feature from few-shot data extracted using $f$. The cost metric is defined as the cosine distance between $i$ and $j$, aligning with our classification criteria in Sec.~\ref{sec:pt_framework}.
$\Omega_{\alpha}$ is a Laplacian regularization term that aims to preserve data structure during transport~\cite{ferradans2014regularized}.

Once the OT plan $\gamma^*$ is obtained, we use it as the transformation function to shift the learned prototypes toward the region where the few-shot data features reside, while preserving their original geometric information. Specifically, each shifted prototype center $\vp^*$ is generated by a barycentric projection: \(\vp^*=\sum^{M_t}_{j=1}\pi_{i,j}\vz_{t,j}\), where \(\pi_{i,j}=\frac{M_{i,j}}{\sum^{M_t}_{j=1}M_{i,j}}\) is the normalized transport plan and \(\vz_{t,j}\) is the latent feature of the $j$-th target-domain data sample. These transformed prototypes \(\mP^*=\{\vp^*_1, \ldots, \vp^*_{2K}\}\) are then deployed as final classifiers. Note that although OT formulation is initially designed for unsupervised transformation~\citep{courty2016optimal}, we perform class-wise transformation based on the label information to retain the discrimination capability of transformed prototypes.

\begin{table*}[t!]
    \centering
    \resizebox{.95\textwidth}{!}{%
        \begin{tabular}{l|ccc|ccc|ccc|ccc|c}
            \toprule
            \multirow{3}{*}{Method}&\multicolumn{3}{c}{\textbf{OCI→M}}&\multicolumn{3}{c}{\textbf{OMI→C}}&\multicolumn{3}{c}{\textbf{OCM→I}}&\multicolumn{3}{c}{\textbf{ICM→O}}&Avg.\\ \cline{2-14} 
            &\multirow{2}{*}{HTER$\downarrow$}&\multirow{2}{*}{AUC$\uparrow$}&\multirow{2}{*}{\begin{tabular}[c]{@{}c@{}}TPR@\\ FPR=1\%$\uparrow$ \end{tabular}}&\multirow{2}{*}{HTER$\downarrow$}&\multirow{2}{*}{AUC$\uparrow$}&\multirow{2}{*}{\begin{tabular}[c]{@{}c@{}}TPR@\\ FPR=1\%$\uparrow$ \end{tabular}}&\multirow{2}{*}{HTER$\downarrow$}&\multirow{2}{*}{AUC$\uparrow$}&\multirow{2}{*}{\begin{tabular}[c]{@{}c@{}}TPR@\\ FPR=1\%$\uparrow$ \end{tabular}}&\multirow{2}{*}{HTER$\downarrow$}&\multirow{2}{*}{AUC$\uparrow$}&\multirow{2}{*}{\begin{tabular}[c]{@{}c@{}}TPR@\\ FPR=1\%$\uparrow$ \end{tabular}}&\multirow{2}{*}{HTER$\downarrow$}\\
            & &&&&&&&&&&&&\\ \midrule
            
            MADDG~\cite{Rui2019Multi}&17.69&88.06&-&24.50&84.51&-&22.19&84.99&-&27.98&80.02&-&23.09\\
            
            DR-MD-Net~\cite{Wangarticle}&17.02&90.10&-&19.68&87.43&-&20.87&86.72&-&25.02&81.47&-&20.64\\
            
            RFMeta~\cite{shao2020regularized}&13.89&93.98&-&20.27&88.16&-&17.30&90.48&-&16.45&91.16&-&16.97\\
            
            NAS-FAS~\cite{yu2020nasfas}&19.53&88.63&-&16.54&90.18&-&14.51&93.84&-&13.80&93.43&-&16.09\\
            
            D$^{2}$AM~\cite{chen2021generalizable}&12.70&95.66&-&20.98&85.58&-&15.43&91.22&-&15.27&90.87&-&16.09\\
            
            SDA~\cite{wang2021self}&15.40&91.80&-&24.50&84.40&-&15.60&90.10&-&23.10&84.30&-&19.65\\
            
            DRDG~\cite{liu2021dual}&12.43&95.81&-&19.05&88.79&-&15.56&91.79&-&15.63&91.75&-&15.66\\
            
            ANRL~\cite{liu2021adaptive}&10.83&96.75&-&17.83&89.26&-&16.03&91.04&-&15.67&91.90&-&15.09\\
            
            SSDG-R~\cite{jia2020single}&7.38&97.17&-&10.44&95.94&-&11.71&96.59&-&15.61&91.54&-&11.28\\
            
            SSAN-R~\cite{wang2022domain}&6.67&98.75&-&10.00&96.67&-&8.88&96.79&-&13.72&93.63&-&9.81\\
            
            PatchNet~\cite{Wang_2022_CVPR}&7.10&98.46&-&11.33&94.58&-&13.40&95.67&-&11.82&95.07&-&10.91\\
            
            SA-FAS~\cite{sun2023rethinking}&5.95&96.55&-&8.78&95.37&-&6.58&97.54&-&10.00&96.23&-&7.82\\
            
            IADG~\cite{zhou2023instance}&5.41&98.19&-&8.70&96.44&-&10.62&94.50&-&8.86&97.14&-&8.39\\

            % added during rebuttal
            GDA~\cite{zhou2022generative} & 9.20 & 98.00 & - & 12.20 & 93.00 & - & 10.00 & 96.00 & - & 14.40 & 92.60 & - & 11.45  \\
            HPDR~\cite{hu2024rethinking} & 4.58 & 96.02 & - & 11.30 & 94.42 & - & 11.26 & 92.49 & - & 9.93 & 95.26 & - & 9.27 \\
            SDA-FAS~\cite{liu2022source} & 5.00 & 97.96 & - & 2.40 & 99.72 & - & 2.62 & 99.48 & - & 5.07 & 99.01 & - & 3.77 \\
            TTDG~\cite{zhou2024test} & 4.16 & 98.48 & - & 7.59 & 98.18 & - & 9.62 & 98.18 & - & 10.00 & 96.15 & - & 7.84 \\

            CFPL~\cite{liu2024cfpl}&3.09&99.45&94.28&2.56&99.10&66.33&5.43&98.41&85.29&3.33&99.05&90.06&3.60\\ 
            
            \rowcolor{backcolour} 
            \textbf{\ours$^*$} (zero-shot)&2.62&99.34&92.38&2.22&99.49&90.67&5.32&98.44&89.00&3.56&99.34&88.64&3.43\\ 

            \midrule
            
            ViTA~\cite{huang2022adaptive} (5-shot) & \transrowcolor 4.75 & \transrowcolor98.84 & \transrowcolor76.67 & \transrowcolor5.00 & \transrowcolor99.13 & \transrowcolor82.14 & \transrowcolor5.37 & \transrowcolor98.57 & \transrowcolor76.15 & \transrowcolor7.16 & \transrowcolor97.97 & \transrowcolor73.24 & \transrowcolor5.57 \\
            
            ViTAF~\cite{huang2022adaptive} (5-shot) & \transrowcolor3.42 & \transrowcolor99.30 & \transrowcolor88.33 & \transrowcolor1.40 & \transrowcolor99.85 & \transrowcolor95.71 & \transrowcolor3.74 & \transrowcolor99.34 & \transrowcolor85.38 & \transrowcolor7.17 & \transrowcolor98.26 & \transrowcolor71.97 & \transrowcolor3.93 \\ 
             
            \midrule
            
            \rowcolor{backcolour}
            \textbf{\ours$^\dagger$} (training-free)&2.38&99.42& 93.33 &2.67&99.49&91.11&5.19&98.56&88.22&3.03&99.45&90.66&3.21\\ 

            \rowcolor{backcolour}
            \textbf{\ours$^\ddagger$} (lightweight)&\textbf{2.14}&\textbf{99.47}&\textbf{95.23}&\textbf{2.00}&\textbf{99.75}&\textbf{93.79}&\textbf{4.85}&\textbf{98.81}&\textbf{91.30}&\textbf{2.61}&\textbf{99.52}&\textbf{92.30}&\textbf{2.91}\\

            \bottomrule
        \end{tabular}
    }
    \caption{The results (\%) of cross-domain evaluation on MSU-MFSD (M), CASIA-FASD (C), ReplayAttack (I), and OULU-NPU (O) datasets. Note that symbols $*$, $\dagger$, and $\ddagger$ indicate three versions of \ours: zero-shot (Section~\ref{sec:pt_framework}), training-free domain adaptation (Section~\ref{sec:ot_adapt}), and lightweight training domain adaptation (Section~\ref{sec:geodesic}), respectively. Baseline results are sourced from \citet{liu2024cfpl}. Both of our approaches outperform existing DG methods and achieve competitive performance against FSL methods (colored in gray). Among our proposed approaches, the lightweight training method delivers the best performance, as it trains with augmented feature data from geodesic mixup to more effectively capture the geometry of the feature space.}
    \label{tab:MCIO_results}
    \vspace{-3mm}
\end{table*}

\subsection{Geodesic Mixup}
\label{sec:geodesic}
In this section, we explore an alternative approach by conducting a lightweight training procedure that uses optimal transport in a different fashion. Specifically, we treat both the prototypes and few-shot target-domain data features as training data in the latent space and learn a separate decision boundary from scratch. It is well-known that classifiers trained on few-shot data can lead to skewed decision boundaries, which in turn result in poor performance~\cite{koch2015siamese, snell2017prototypical}. A common remedy for this issue involves generating synthetic data as augmentations~\cite{devries2017improved, zhang2018metagan, shafahi2019adversarial}. Among these methods, mixup~\cite{zhang2017mixup, verma2019manifold} is a simple option that can also be applied in the embedding space. Nevertheless, such instance-wise interpolation does not account for global information (\textit{i.e.}, geometry) of a distribution, which is critical for our scenario, generating inferior results due to unexpected artifacts~\cite{lamb2019interpolated}.

To address this challenge, we propose leveraging discrete Wasserstein barycenters with free support~\cite{alvarez2016fixed} to generate synthesized data. In our approach, the problem involves only two distributions, empirically represented by prototypes from the source domains and few-shot data features from the target domain.
Searching for barycenters is equivalent to interpolating along the Wasserstein geodesic between these source and target distributions. Concretely, given a mixing coefficient $w \in [0,1]$, we look for a distribution \(\mu\) that minimizes its Wasserstein distance to both the source and target distributions:
\begin{equation}
   \mu = \min_{\mu} \bigl[w W(\mu, \mu_{s}) + (1-w) W(\mu, \mu_{t})\bigr],
\label{eq:barycenter}
\end{equation}
where $W(\cdot, \cdot)$ denotes the Wasserstein distance between two distributions.
In our case, \(\mu_{s}\) is substantiated by learned source prototypes and \(\mu_{t}\) contains few-shot target-domain features. \(\mu\) is modeled by $Q$ equally important supports: \(\mu = \sum_{i=1}^{Q} \frac{1}{Q} \delta_{x_i}\), where \(\delta_{x_i}\) refers to the Dirac function at position $x_i$. In practice, we set $Q$ to be $K$, the total number of sub-centers within the learned prototypes for either bona fide or spoof. We then generate \(\mu\) on the fly at each iteration as an augmented data batch according to a randomly sampled $w$ from a beta distribution: $w \sim \text{Beta}(0.4, 0.4)$. Unlike point-wise interpolation (\textit{i.e.}, standard mixup strategies), using generated \(\mu\) respects the geometries of both source and target distributions, thus leading to a better discrimination boundary.

In implementation, we solve the entropy-regularized version of Eq.~\ref{eq:barycenter} that supports the efficient Sinkhorn algorithm~\cite{sinkhorn1967diagonal} and reduces sparsity. To further encourage vicinity risk minimization, we also perturb few-shot data features at each iteration according to common data augmentation techniques (\textit{e.g.}, color jittering, random Gaussian noise).

\section{Experimental Results}
In this section, we compare \ours with state-of-the-art FAS methods using the widely adopted cross-domain and cross-attack evaluation protocol. To further validate the effectiveness of \ours, we benchmark it against several test-time few-shot adaptation methods from general domains. Additionally, we conduct a series of ablation studies to evaluate the functionality of each component in \ours.

\subsection{Experiment Settings}
\noindent\textbf{Datasets.} We evaluate our method on four standard benchmarks: \texttt{Idiap Replay Attack} \cite{chingovska2012indiap_replay} (\textbf{I}), \texttt{OULU-NPU} \cite{boulkenafet2017oulu_npu} (\textbf{O}), \texttt{CASIA-MFSD} \cite{zhang2012casia} (\textbf{C}), and \texttt{MSU-MFSD} \cite{wen2015msu_mfsd} (\textbf{M}). Following the approach of previous works, we treat each dataset as a distinct domain and employ a leave-one-out testing protocol to evaluate cross-domain performance. For example, the protocol \textbf{OCI $\rightarrow$ M} indicates training on \texttt{OULU-NPU}, \texttt{CASIA-MFSD}, and \texttt{Idiap Replay Attack}, and testing on \texttt{MSU-MFSD}.

\noindent\textbf{Evaluation Metrics.} We use three commonly adopted metrics to quantify performance: Half Total Error Rate (HTER), Area Under the Receiver Operating Characteristic Curve (AUC), and True Positive Rate (TPR) at a False Positive Rate (FPR) of 1\% (TPR@FPR=1\%).

\noindent\textbf{Implementation Details.} We initialize the feature extractor using the pretrained ViT-B/16 model from CLIP~\cite{radford2021learning}. The number of sub-centroids $K$ for each class is set to 50 by default, and the strength scalars $\alpha$, $\beta$, and $\eta$ are set to $0.01$, $0.01$, and $1.0$, respectively. At the test stage, we use 10 randomly selected images per class from the target domain as few-shot examples, and the strength $\lambda$ of the Laplacian regularization term is fixed at 100. For learning a separate decision boundary, we optimize a linear classifier for 100 iterations using standard cross-entropy loss and the Adam optimizer~\cite{kingma2014adam} with a learning rate of 0.01.

\begin{figure}[t]
\begin{minipage}{0.24\textwidth}
\centering
\resizebox{.9\textwidth}{!}{%
\begin{tabular}{l|c}
	\toprule
	Methods & AUC$\uparrow$   \\
	\midrule
	SVM1+IMQ~\cite{arashloo2017anomaly} & 70.23$^{\text{12.69}}$ \\
	% CDCN~\cite{yu2020searching} & {88.69}$^{\text{10.56}}$  \\
	CDCN++~\cite{yu2020searching} & {87.53}$^{\text{10.90}}$ \\
	SSAN~\cite{wang2022ssan} & {88.01}$^{\text{9.93}}$ \\
	TTN-S~\cite{wang2022learning} &{89.71}$^{\text{9.17}}$ \\
	UDG-FAS~\cite{liu2023udgfasssdg} & {92.43}$^{\text{6.86}}$ \\
        GAC-FAS~\cite{le2024gradient} &  93.39$^{\text{4.27}}$ \\
    \hline
    \rowcolor{backcolour} \textbf{\ours$^*$} (zero-shot) &    \textbf{98.32}$^{\text{0.24}}$ \\
    \rowcolor{backcolour} \textbf{\ours$^\dagger$} (training-free) &    \textbf{98.38}$^{\text{0.10}}$ \\
    \rowcolor{backcolour} \textbf{\ours$^\ddagger$} (lightweight) &    \textbf{98.54}$^{\text{0.29}}$ \\
    \bottomrule
\end{tabular}%
}
\subcaption{Unseen 2D attack}
\end{minipage}%
\begin{minipage}{0.24\textwidth}
\centering
\resizebox{.9\textwidth}{!}{%
\begin{tabular}{l|l}
    \toprule
    Method & AUC$\uparrow$   \\
    \midrule
    Saha \textit{et al.} \cite{saha2020domain} & 79.20 \\
    Panwar \textit{et al.} \cite{panwar2021unsupervised} & 80.00 \\
    SSDG-R~\cite{jia2020ssgd} & 82.11 \\
    CIFAS~\cite{liu2022causal} & 83.20 \\
    UDG-FAS~\cite{liu2023udgfasssdg} & 87.26 \\
    GAC-FAS~\cite{le2024gradient} & 89.27$^{\text{0.58}}$ \\
    \hline
    \rowcolor{backcolour} \textbf{\ours$^*$} (zero-shot) &    \textbf{98.62}$^{\text{0.52}}$ \\
    \rowcolor{backcolour} \textbf{\ours$^\dagger$} (training-free) &    \textbf{98.75}$^{\text{0.79}}$ \\
    \rowcolor{backcolour} \textbf{\ours$^\ddagger$} (lightweight) &    \textbf{99.63}$^{\text{0.11}}$ \\
    \bottomrule
\end{tabular}%
}
\subcaption{Unseen 3D attack}
\end{minipage}
\captionof{table}{{Evaluation on cross-attack protocol following \citet{arashloo2017anomaly}}. Baseline results are sourced from \citet{le2024gradient}.}
\addtocounter{figure}{-1}
\label{tab:unseen}
\vspace{-5mm}
\end{figure}

\begin{table*}[t]
\centering
\resizebox{.99\textwidth}{!}{%
    \begin{tabular}{lccccccccccccc}
    \toprule
    \multirow{3}{*}{Method}&\multicolumn{3}{c}{\textbf{OCI→M}}&\multicolumn{3}{c}{\textbf{OMI→C}}&\multicolumn{3}{c}{\textbf{OCM→I}}&\multicolumn{3}{c}{\textbf{ICM→O}}&Avg.\\ \cline{2-14} 
    &\multirow{2}{*}{HTER$\downarrow$}&\multirow{2}{*}{AUC$\uparrow$}&\multirow{2}{*}{\begin{tabular}[c]{@{}c@{}}TPR@\\ FPR=1\%$\uparrow$ \end{tabular}}&\multirow{2}{*}{HTER$\downarrow$}&\multirow{2}{*}{AUC$\uparrow$}&\multirow{2}{*}{\begin{tabular}[c]{@{}c@{}}TPR@\\ FPR=1\%$\uparrow$ \end{tabular}}&\multirow{2}{*}{HTER$\downarrow$}&\multirow{2}{*}{AUC$\uparrow$}&\multirow{2}{*}{\begin{tabular}[c]{@{}c@{}}TPR@\\ FPR=1\%$\uparrow$ \end{tabular}}&\multirow{2}{*}{HTER$\downarrow$}&\multirow{2}{*}{AUC$\uparrow$}&\multirow{2}{*}{\begin{tabular}[c]{@{}c@{}}TPR@\\ FPR=1\%$\uparrow$ \end{tabular}}&\multirow{2}{*}{HTER$\downarrow$}\\
    & &&&&&&&&&&&&\\ \midrule

    NCM~\cite{mai2021supervised} & 3.25 & 98.16 & 29.68 & 3.15 & 98.23 & 31.48 & 7.97 & 96.60 & 31.57 & 3.28 & 99.43 & 89.03 & 4.41 \\
    Tip-Adaptor~\cite{zhang2021tip} & 3.73 & 98.87 & 85.08 & 3.37 & 99.28 & 89.83 & 9.28 & 93.44 & 29.53 & 3.05 & \textbf{99.46} & 88.87 & 4.84 \\

    \rowcolor{backcolour}
    \textbf{\ours$^\dagger$} (training-free)&\textbf{2.38}&\textbf{99.42}&\textbf{93.33}&\textbf{2.67}&\textbf{99.49}&\textbf{91.11}&\textbf{5.19}&\textbf{98.56}&\textbf{88.22}&\textbf{3.03}&99.45&\textbf{90.66}&\textbf{3.21}\\ 
    
    \midrule
    \midrule
    
    Linear Probe~\cite{radford2021learning} & 2.14 & 99.46 & 78.57 & 2.22 & 99.64 & 92.89 & 5.38 & 98.74 & 88.00 & 3.12 & 99.41 & 90.80 & 3.22 \\
    
    Manifold Mixup~\cite{verma2019manifold} & 2.14 & 99.45 & 69.05 & 2.16 & 99.63 & 93.10 & 5.32 & 98.80 & 90.1 & 2.82 & 99.42 & 91.19 & 3.04 \\
    \rowcolor{backcolour}
    \textbf{\ours$^\ddagger$} (lightweight)&\textbf{2.14}&\textbf{99.47}& \textbf{95.23}& \textbf{2.00} &\textbf{99.75}& \textbf{93.79}&\textbf{4.85}&\textbf{98.81}&\textbf{91.30}&\textbf{2.61}&\textbf{99.52}&\textbf{92.30}&\textbf{2.91}\\ 
    \bottomrule
    \end{tabular}
    }
\caption{Comparison to few-shot adaptation methods. We compare \ours against several representative few-shot adaptation methods in the FAS context. Our method achieves overall better performance in HTER.} 
\label{tab:fewshot_adapt}
\vspace{-2mm}
\end{table*}

\begin{table*}[]
\centering
\resizebox{.99\textwidth}{!}{%
    \begin{tabular}{lccccccccccccc}
    \toprule
    \multirow{3}{*}{Method}&\multicolumn{3}{c}{\textbf{OCI→M}}&\multicolumn{3}{c}{\textbf{OMI→C}}&\multicolumn{3}{c}{\textbf{OCM→I}}&\multicolumn{3}{c}{\textbf{ICM→O}}&Avg.\\ \cline{2-14} 
    &\multirow{2}{*}{HTER$\downarrow$}&\multirow{2}{*}{AUC$\uparrow$}&\multirow{2}{*}{\begin{tabular}[c]{@{}c@{}}TPR@\\ FPR=1\%$\uparrow$ \end{tabular}}&\multirow{2}{*}{HTER$\downarrow$}&\multirow{2}{*}{AUC$\uparrow$}&\multirow{2}{*}{\begin{tabular}[c]{@{}c@{}}TPR@\\ FPR=1\%$\uparrow$ \end{tabular}}&\multirow{2}{*}{HTER$\downarrow$}&\multirow{2}{*}{AUC$\uparrow$}&\multirow{2}{*}{\begin{tabular}[c]{@{}c@{}}TPR@\\ FPR=1\%$\uparrow$ \end{tabular}}&\multirow{2}{*}{HTER$\downarrow$}&\multirow{2}{*}{AUC$\uparrow$}&\multirow{2}{*}{\begin{tabular}[c]{@{}c@{}}TPR@\\ FPR=1\%$\uparrow$ \end{tabular}}&\multirow{2}{*}{HTER$\downarrow$}\\
    & &&&&&&&&&&&&\\  \midrule
     SSDG-R \cite{jia2020ssgd}   & 22.84$^{1.14}$  & 78.67$^{1.31}$  &- & 28.76$^{0.89}$ & 80.91$^{1.10}$  & - & 14.65$^{1.21}$ & 91.93$^{1.35}$  & - & 15.83$^{1.29}$ & 92.13$^{0.96}$ & - & 20.52\\
     SSAN-R  \cite{wang2022ssan}  & 21.79$^{3.68}$ & 84.06$^{3.78}$ & - & 26.44$^{2.91}$ & 78.84$^{2.83}$ & - & 35.39$^{8.04}$ & 70.13$^{9.03}$ & - & 25.72$^{3.74}$ & 79.37$^{4.69}$ & -& 27.34 \\
     PatchNet \cite{wang2022patchnet}  & 25.92$^{1.13}$ & 83.43$^{0.87}$ & - & 36.26$^{1.98}$ & 71.38$^{1.89}$ & - & 29.75$^{2.76}$ & 80.53$^{1.35}$ & - & 23.49$^{1.80}$ & 84.62$^{1.92}$ &- & 28.86\\
     SA-FAS \cite{sun2023safas} & 14.36$^{1.10}$ & 92.06$^{0.53}$ & - & 19.40$^{0.66}$ & 88.69$^{0.67}$ & - & 11.48$^{1.10}$ & 95.74$^{0.55}$ & - & 11.29$^{0.32}$ & 95.23$^{0.24}$ & - & 14.13\\
     GAC-FAS~\cite{le2024gradient} & 12.29$^{1.29}$ & 95.35$^{0.57}$ & - & 15.37$^{1.52}$ & 91.67$^{1.67}$ & - & 12.51$^{3.03}$ & {93.03}$^{2.24}$ & - & 9.89$^{0.47}$ & 96.44$^{0.18}$ & - & 12.52 \\
    \rowcolor{backcolour} 
    \textbf{\ours$^*$}(zero-shot)&\textbf{4.52}$^{0.70}$&\textbf{99.18}$^{0.25}$&\textbf{86.19}$^{5.10}$&\textbf{4.33}$^{0.89}$&\textbf{99.23}$^{0.20}$&\textbf{88.00}$^{1.75}$&\textbf{7.98}$^{1.33}$&\textbf{96.64}$^{0.67}$&\textbf{84.97}$^{1.33}$&\textbf{4.36}$^{0.15}$&\textbf{99.09}$^{0.09}$&\textbf{86.50}$^{1.71}$&\textbf{5.29}\\ 
    \bottomrule
    \end{tabular}
}
\caption{{Evaluation at convergence.} Following the evaluation method proposed in \cite{sun2023safas}, we compare the zero-shot version of \ours with baseline methods upon convergence. The baseline results are sourced from \cite{le2024gradient}. \ours consistently exhibits superior convergence performance measured by HTER, AUC, and TPR@FPR=1\%.}
\label{tab:convergence}
\vspace{-5mm}
\end{table*}

\subsection{Comparison to SoTA methods}

\noindent\textbf{Cross-domain Evaluation.}
Table~\ref{tab:MCIO_results} presents the cross-domain performance of \ours under a comprehensive evaluation protocol. As shown, although our primary focus is not on improving domain generalization (DG) performance, the proposed prototype-based framework achieves state-of-the-art (SoTA) DG performance (indicated by $*$), demonstrating the effectiveness of our proposed representation learning described in Section~\ref{sec:pt_framework}. Without additional training, \ours improves performance on the target domain by leveraging few-shot examples (denoted by $\dagger$), particularly excelling in the most challenging metric, TPR@FPR=1\%. Our lightweight training with geodesic mixup (indicated by $\ddagger$) improves performance further, surpassing existing methods by a clear margin. Unlike previous few-shot approaches in the FAS domain, such as ViTAF~\cite{huang2022adaptive}, which assumes that few-shot data from the target domain is available during the training phase, our approach treats few-shot data as examples that can only be incorporated at the test stage. This makes \ours a more practical yet challenging setting.

\noindent\textbf{Cross-attack Evaluation.} In real-world applications, security models must contend with increasingly sophisticated attack methods. Evaluating \ours against unseen attacks is critical to validate its robustness. Following the protocol proposed in \cite{arashloo2017anomaly}, we simulate unseen 2D attacks by training on two domains from \textbf{I}, \textbf{C}, and \textbf{M}, and testing on an unseen attack from an unseen domain. For 3D attacks, we train on \textbf{O}, \textbf{C}, and \textbf{M}, and evaluate using a subset from the CelebA-Spoof dataset~\cite{zhang2020celeba}. As shown in Table~\ref{tab:unseen}, \ours exhibits strong generalization performance when confronted with unseen attacks. Additionally, both the training-free and lightweight adaptation variants further improve performance by incorporating few-shot examples of the new attack during test time.

\begin{table*}[t]
\centering
\resizebox{.99\textwidth}{!}{%
    \begin{tabular}{lcccccccccccc}
    \toprule
    \multirow{3}{*}{Method}&\multicolumn{3}{c}{\textbf{OCI→M}}&\multicolumn{3}{c}{\textbf{OMI→C}}&\multicolumn{3}{c}{\textbf{OCM→I}}&\multicolumn{3}{c}{\textbf{ICM→O}}\\ \cline{2-13} 
    &\multirow{2}{*}{HTER$\downarrow$}&\multirow{2}{*}{AUC$\uparrow$}&\multirow{2}{*}{\begin{tabular}[c]{@{}c@{}}TPR@\\ FPR=1\%$\uparrow$ \end{tabular}}&\multirow{2}{*}{HTER$\downarrow$}&\multirow{2}{*}{AUC$\uparrow$}&\multirow{2}{*}{\begin{tabular}[c]{@{}c@{}}TPR@\\ FPR=1\%$\uparrow$ \end{tabular}}&\multirow{2}{*}{HTER$\downarrow$}&\multirow{2}{*}{AUC$\uparrow$}&\multirow{2}{*}{\begin{tabular}[c]{@{}c@{}}TPR@\\ FPR=1\%$\uparrow$ \end{tabular}}&\multirow{2}{*}{HTER$\downarrow$}&\multirow{2}{*}{AUC$\uparrow$}&\multirow{2}{*}{\begin{tabular}[c]{@{}c@{}}TPR@\\ FPR=1\%$\uparrow$ \end{tabular}}\\
    & &&&&&&&&&&&\\ \midrule     
    
    \textbf{\ours} (DG) &2.62&99.34&90.49&2.22&99.49&90.67&\bf5.32&\bf98.44&\bf88.13&3.56&99.34&88.64\\ 

    \textbf{\ours} (spoof only)&2.38&99.40&90.49&2.33&99.49&\textbf{91.11}&5.83&98.24&88.00&3.53&99.35&88.41\\
    
    \textbf{\ours} (bona fide only) &\textbf{2.38}&\textbf{99.42}&\textbf{92.87}&\textbf{2.22}&\textbf{99.53}&90.89&6.42&98.29&85.50&\textbf{3.38}&\textbf{99.36}&\textbf{89.65} \\ 
    \midrule
    
    \textbf{\ours} (both) &2.38&99.42&93.33&2.67&99.49&91.11&5.19&98.56& 88.22&3.03&99.45&90.66 \\ 
    
    \bottomrule
    \end{tabular}
    }
\caption{\textbf{Empirical results of \ours under the one-class setting.} Providing few-shot data from the bona fide class proves to be more effective than from the spoofing class in most scenarios. One exception is OCM→I, where I focuses exclusively on replay attacks under controlled environments. Consequently, both the bona fide and spoof data of \textbf{I} differ markedly from those of the other three datasets, and \ours needs both bona fide and spoof data to achieve satisfactory adaptation results.}
\label{tab:one_class}
\vspace{-4mm}
\end{table*}

\begin{table}[t]
	\centering
	\resizebox{1.0\linewidth}{!}{
		\begin{tabular}{@{}cccc|ccc@{}}
		\toprule
		\textbf{proto.} & \textbf{orth.} & \textbf{coarse con.} & \textbf{fine con.} & HTER(\%)$\downarrow$ & AUC(\%)$\uparrow$  & \begin{tabular}[c]{@{}c@{}}TPR(\%) $\uparrow$\\ @FPR=1\% \end{tabular} \\ 
            \midrule
            $\checkmark$  & -  & - & -  & 5.20 & 98.37 & 81.34 \\
            $\checkmark$  & $\checkmark$  & - & -  & 5.14 & 98.38 & 82.68 \\
            $\checkmark$ & $\checkmark$ & $\checkmark$ & - & 4.87 & 98.51 & 84.81 \\
            $\checkmark$ & $\checkmark$ & $\checkmark$ & $\checkmark$ & \textbf{4.19} & \textbf{98.62} & \textbf{87.35} \\
            \bottomrule
		\end{tabular}
	}
	\caption{Ablation of learning objectives introduced in Section~\ref{sec:pt_framework}. Each loss term effectively improves the DG performance of our prototype-based framework.}
	\label{tab:ab_loss}
 \vspace{-5mm}
\end{table}

\subsection{Comparison to Few-shot Adaptation Methods}
\label{sec:fs}

In this section, we examine the effectiveness of \ours by comparing it with other comparable test-time few-shot adaptation methods. Specifically, we implement the following four baseline methods, covering both training-free and lightweight adaptation schemes.

\noindent\textbf{Nearest Class Mean (NCM)}~\cite{mai2021supervised}. We implement NCM using only class mean feature from few-shot data as the classifier. As shown in Table~\ref{tab:fewshot_adapt}, NCM lags behind \ours by a significant margin, highlighting the importance of knowledge transfer from source domains under few-shot conditions.

\noindent\textbf{Tip-Adaptor}~\cite{zhang2021tip}. We compare \ours with the state-of-the-art training-free few-shot adaptation method. Tip-Adaptor relies on an additional validation set for hyperparameter selection. Hence, we build a validation set with 256 randomly selected images from existing source domains. As shown in Table~\ref{tab:fewshot_adapt}, \ours demonstrates a clear advantage over Tip-Adaptor and even matches its performance when a large additional validation set is provided for Tip-Adaptor.

\noindent\textbf{Linear Probe}~\cite{radford2021learning}. Training a linear probe over a frozen feature extractor is a simple yet competitive method~\cite{huang2024lp++} when lightweight training is allowed. Here, we train a linear probe over a mixture of our learned prototypes and few-shot features. As shown in Table~\ref{tab:fewshot_adapt}, linear probes achieve strong performance, outperforming other training-free adaptation methods. However, \ours demonstrates a noticeable improvement over a simple linear probe.
    
\noindent\textbf{Manifold Mixup}~\cite{verma2019manifold}. We compare \ours with manifold mixup, which also performs latent space augmentation to boost performance over a linear probe. However, unlike this instance-wise augmentation, our proposed geodesic mixup generates distribution-wise augmentations, which is more effective under our setup.

\subsection{Ablation Studies}

\noindent\textbf{Performance Upon Convergence.} A recent study~\cite{sun2023rethinking} points out that a single snapshot of performance on a test set may not accurately reflect a detection model's generalization capability. Following their approach, we report the average performance of our model across the last 10 evaluations in Table~\ref{tab:convergence}. \ours consistently exhibits superior convergence performance measured by HTER, AUC, and TPR@FPR=1\%, while showing significantly less fluctuation than other methods. For example, as shown in Table~\ref{tab:MCIO_results}, the converged HTER score of SA-FAS increases from 7.82 to 14.13. In contrast, \ours achieves HTER scores of 3.43 at the snapshot and 5.29 at convergence.

\noindent\textbf{Effectiveness of Learning Objectives.} Our proposed prototype-based framework combines a prototype-based margin loss, coarse-to-fine supervised contrastive loss, and an orthogonal loss for supervision. Using the cross-attack (3D) protocol as an example, we show in Table~\ref{tab:ab_loss} the effectiveness of each proposed loss component. Each term effectively improves the DG performance of our prototype-based framework.

\noindent\textbf{Sub-centroid Number.} To increase the expressiveness of prototypes and better capture the geometric information of source domains, we employ a multi-centroid strategy. Using the cross-attack (3D) protocol as an example, we run experiments with the number of sub-centers $K \in \{1, 5, 10, 30, 50\}$, yielding HTER scores of $\{9.40, 6.33, 5.21, 5.07, 5.20\}$, respectively. Increasing the number of centroids effectively improves domain generalization (DG) performance, and the improvement plateaus around 50 sub-centers; we observed similar trends in other experiments. Consequently, we set $K=50$ for all subsequent experiments.

\begin{figure}[t]
  \centering
   \includegraphics[width=\linewidth]{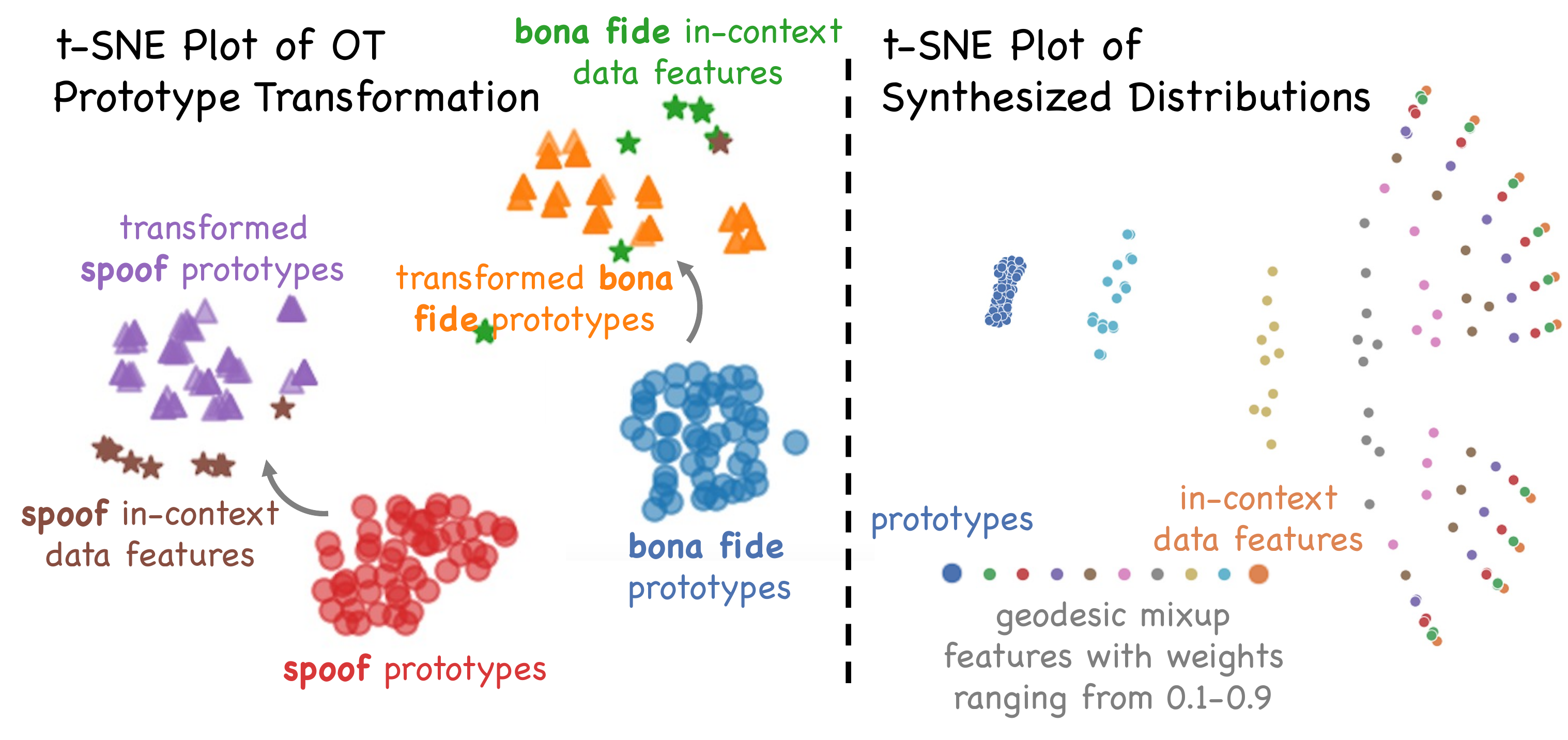}
   \caption{(L) The transformed prototypes are relocated closer to the regions where the few-shot client data features reside, while preserving the geometric information of the original prototypes learned from the source domains. (R) Synthesized distributions are gradually transformed from the source prototypes toward the target few-shot data features, as weight $w$ is varied from 0.1 to 0.9.}
   \vspace{-5mm}
   \label{fig:visualization}
\end{figure}

\noindent\textbf{Visualizations.} We visualize the transformed prototypes and synthesized distributions generated by geodesic mixup in the latent space to provide a clear illustration of our method. As shown in Fig.~\ref{fig:visualization} (L), the transformed prototypes are relocated closer to the regions where the few-shot client data features reside, while preserving the geometric information of the original prototypes learned from the source domains. For geodesic mixup, Fig.~\ref{fig:visualization} (R) shows that synthesized distributions are gradually transformed from the source distribution toward the target distribution as we vary the weight $w$ from 0.1 to 0.9.

\noindent\textbf{Efficiency.} The training-free variant of \ours introduces zero additional learnable parameters and requires a minimal adaptation time of only $0.17\pm0.03$ minutes, estimated over 10 trials. For lightweight adaptation with geodesic mixup, the additional parameter load is 3.9 KB, and the adaptation process completes in $22.73 \pm 3.51$ minutes, also estimated over 10 trials. Our customizable layer offers maximum flexibility and convenience for customization and maintenance by both the host and clients.

\subsection{One-class Scenario}
We extend the training-free version of our approach to the one-class setting, where only a few data points from either the bona fide or spoof class are provided by the client. We transform only the prototypes corresponding to the class that has associated few-shot data, while keeping the other class’s prototypes unchanged. As shown in Table~\ref{tab:one_class}, providing few-shot data from either the bona fide or spoof category helps improve performance compared to DG in most cases. Interestingly, under the one-class setting, providing few-shot data from the bona fide class appears to be more effective than from the spoofing class. We hypothesize that this is because the bona fide class is inherently more compact, so fewer data samples can better represent the target bona fide distribution.
\section{Conclusion}

In this work, we explore a practical yet under-explored scenario in the face anti-spoofing literature, where only a limited set of labeled target-domain data is available at the test stage, and the client portal has no access to the host model. To address this challenge, we propose a prototype-based backbone model, on top of which two efficient adaptation modules are crafted for training-free and lightweight training settings. In particular, we introduce geodesic mixup, an optimal transport-guided synthesis method that generates pseudo empirical distributions as augmentations, thereby improving the learned decision boundary. Extensive empirical results on four benchmarks show that our prototype-based framework achieves performance on par with state-of-the-art domain generalization methods. Furthermore, its performance can be boosted by incorporating few-shot target-domain data at test time, leveraging either the proposed training-free or lightweight training modules. Our approach, which can be integrated with other face recognition algorithms, establishes a simple yet strong baseline for future research.

\newpage
\clearpage
\setcounter{page}{1}
\maketitlesupplementary

\section{Applications of Our Settings}
It is challenging to build a generalized training set that adequately covers all possible environmental conditions, user behaviors, demographic diversity, and specific clients' requirements that may sometimes conflict with one another.
For example, consider a scenario where Client A authenticates users through mobile phone-based checks, while Client B uses entrance kiosks at fixed locations to capture data from specific angles. These diverse setups introduce variations in the data capture process, including acquisition devices, pre-processing techniques, and challenge-response mechanisms.
As a result, there is a domain mismatch between the client data and the training data used to develop the host model.
This mismatch can be catastrophic for face anti-spoofing models, as they are particularly sensitive to low-level image features such as task irrelevant noise and artifacts.
In our work, we aim to address the above-mentioned problem by building a privileged system that allows for lightweight customization at testing stage by either the host or the client, using only a few labeled samples provided by the client.

\noindent\textbf{Relation to Classical Domain Adaptation.}: Classical domain adaptation typically requires explicit access to the source domain data to align the source and target domains. However, in many practical scenarios, access to the source data may be restricted, especially for data involving facial images. Additionally, since the application is related to security, host model parameters cannot be explicitly shared with clients or end-users due to the risks of model theft or white-box adversarial attacks. Therefore, in our setting, both the host model parameters and source training data are not accessible. Instead, only a few data features and prototypes are made available to clients, allowing them to improve the performance for their specific use cases.

\section{Interpretation of Geodesic Mix-up}
We use optimal transport (OT) to compute intermediate distributions along the geodesic path between the source and target data distributions, which are empirically represented by source prototypes and few-shot client data, respectively. This geodesic path minimizes the cost of transporting probability mass while maintaining a smooth transition between the two distributions. Each intermediate distribution along this path represents a weighted blend of source and target characteristics.
By sampling from these intermediate distributions, we generate data that can be interpreted as geodesic mix-up. Geodesic mix-up extends the traditional mix-up concept by shifting from the feature space to the space of probability distributions.
More specifically, instead of interpolating between individual data points, it interpolates between entire distributions, offering the following two advantages:
\begin{itemize}
\item The \textbf{relationships} among real data points can be effectively preserved to the synthetic data points, as geodesic mixup respects the geometric structure of data distributions.
\item Interpolating at the distribution level reduces the impact of noise and outliers in individual data points.
\end{itemize}
Training on these samples allows the classifier to learn how features transition between domains, so that it can adapt to target-specific characteristics while maintaining its knowledge of the source domain.

Solving Equation~5 will get an intermediate distribution along the geodesic path between the source and target distributions. This intermediate distribution represents a weighted blend of source and target characteristics. The parameter $w \in [0,1]$ in the equation determines the weight of the blend: $w$ close to 0 results in a distribution closer to the source, and $w$ close to 1 results in a distribution closer to the target. Intuitively, Equation~5 aims to find a distribution that transitions smoothly between the source and target by minimizing the transportation cost. If we draw an analogy to Euclidean space, the source, target, and intermediate distributions can be thought of as points, and the intermediate distribution lies on the "straight line" connecting the source and target. However, in Wasserstein space, this "straight line" corresponds to the geodesic path, which captures the optimal transport relationship between the distributions, respecting their geometric structure and preserving relationships among data points.

\begin{figure*}[t!]
    \centering
    \includegraphics[width=0.24\textwidth]{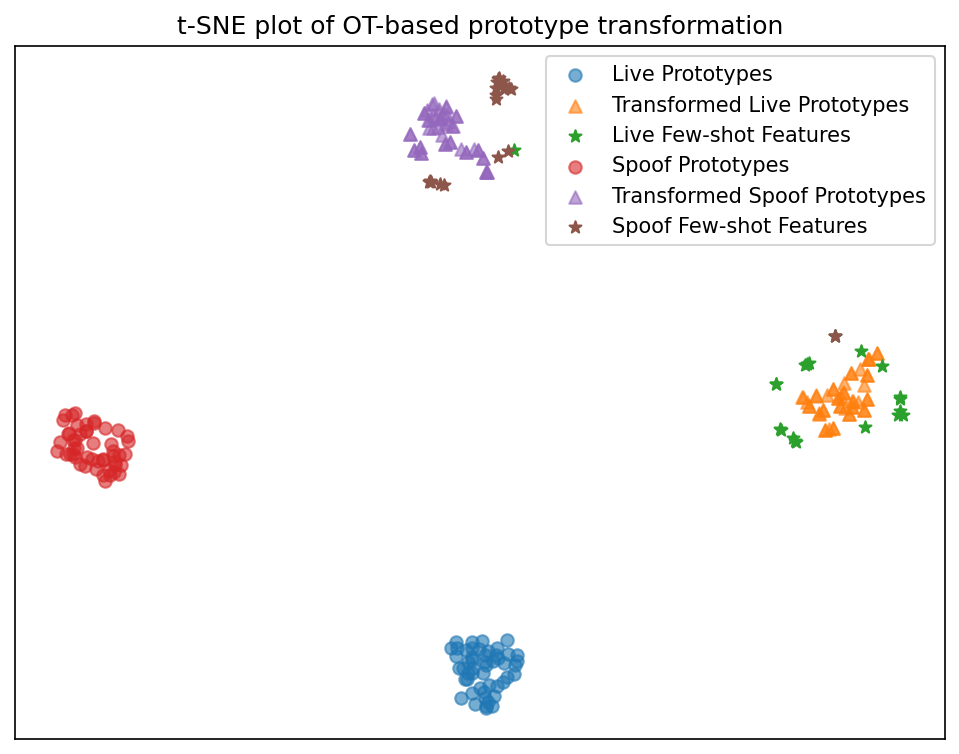}%
    \hfill%
    \includegraphics[width=0.24\textwidth]{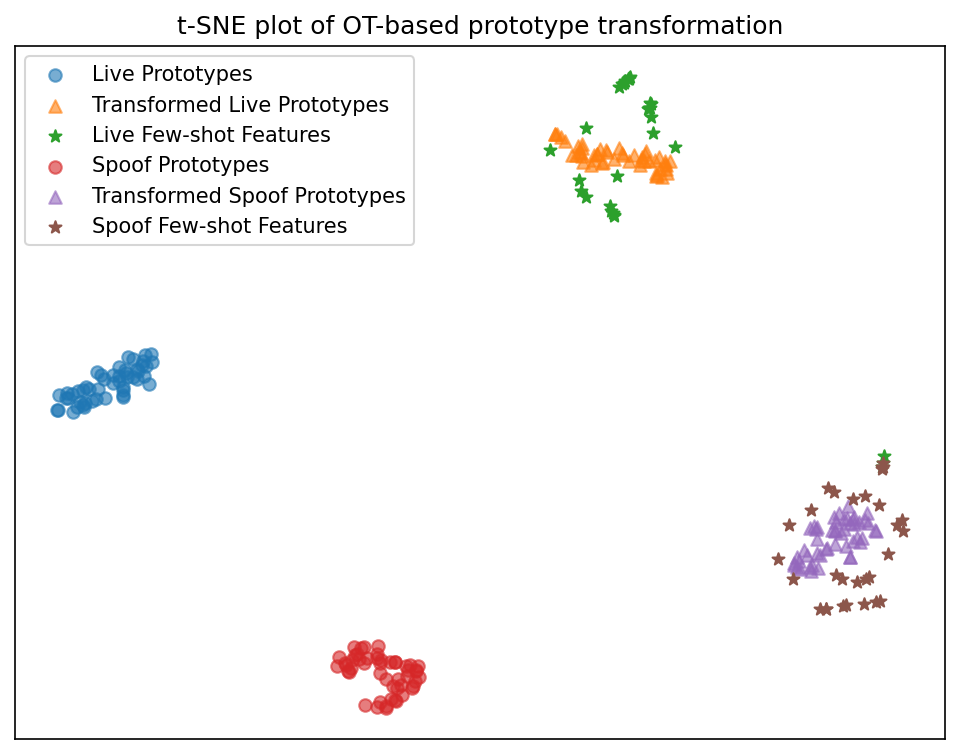}%
    \hfill%
    \includegraphics[width=0.24\textwidth]{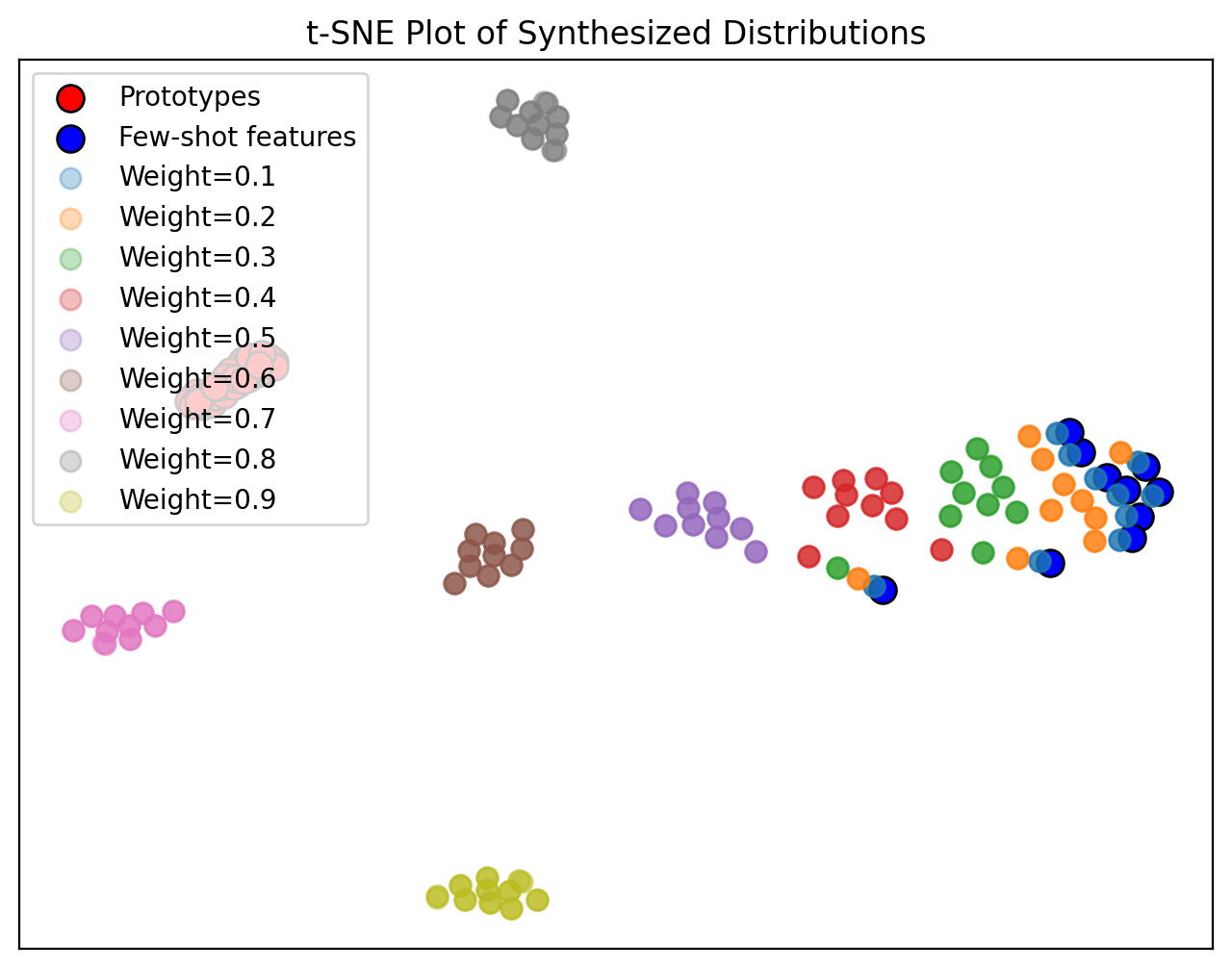}%
    \hfill%
    \includegraphics[width=0.24\textwidth]{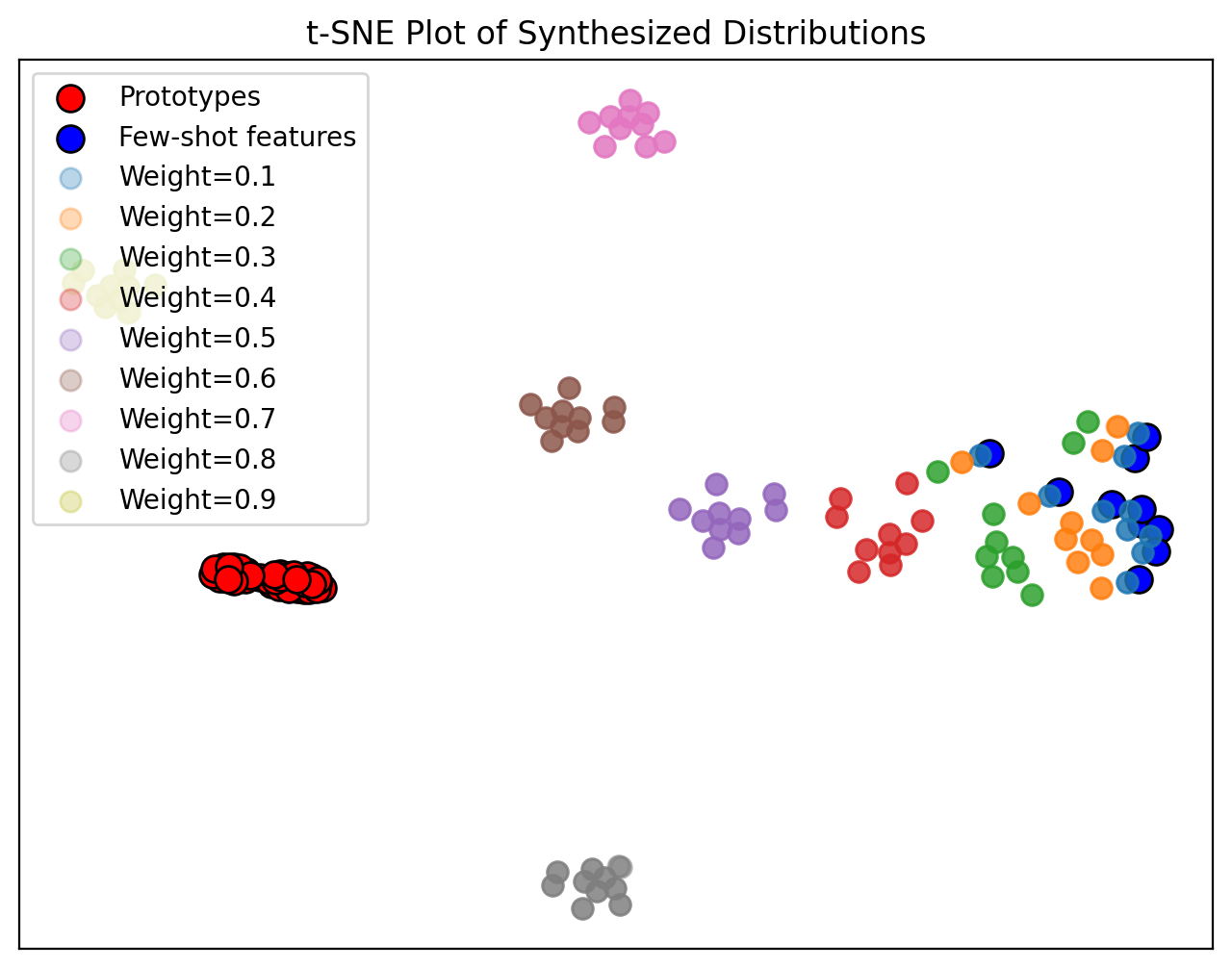}
    \caption{Visualization of OTA in the latent space. Left two plots indicating training-free adaptation. Right two plots resemble the generated synthetic empirical distributions of Geodesic Mixup.}
    \label{fig:ota_viusal}
\end{figure*}

\begin{table*}[t!]
    \centering
    \resizebox{0.95\textwidth}{!}
    {%
        \begin{tabular}{lcccccccccc}
            \toprule
            \multirow{2}{*}{} & \multicolumn{2}{c}{\textbf{K=5}} & \multicolumn{2}{c}{\textbf{K=10}} & \multicolumn{2}{c}{\textbf{K=20}} & \multicolumn{2}{c}{\textbf{K=50}}\\ \cline{2-9}
            & Avg. HTER$\downarrow$ & Avg. AUC$\uparrow$ & Avg. HTER$\downarrow$ & Avg. AUC$\uparrow$ & Avg. HTER$\downarrow$ & Avg. AUC$\uparrow$ & Avg. HTER$\downarrow$ & Avg. AUC$\uparrow$ \\ 
            \midrule
            \textbf{\ours$^\dagger$} (training-free) & 3.57 & 99.05 & 3.21 & 99.23 & 3.13 & 99.27 & 3.01 & 99.30 & \\
            \textbf{\ours$^\ddagger$} (lightweight) & \textbf{3.28} & \textbf{99.29} & \textbf{2.91} & \textbf{99.39} & \textbf{2.82} & \textbf{99.48} & \bf 1.97 & \bf 99.65  \\
            \bottomrule
        \end{tabular}
    }
    \caption{Results of OTA under different few-shot number K.}
    \label{tab:scaling}
\end{table*}

\begin{figure*}[t!]
  \centering
  \vspace{-15pt}
  \resizebox{1.0\textwidth}{!}{
    \begin{minipage}{\textwidth}
      \vspace*{\fill}
      \begin{minipage}[t]{0.49\textwidth}  % Added [t] for top alignment
        \begin{algorithm}[H]
          \begin{algorithmic}[1]
            \Require Source domains $\{\mathcal{D}_i\}_{i=1}^N$, \\
            Feature extractor $f : \mathcal{X} \rightarrow \mathbb{R}^D$, \\ Target support set $\mathcal{D}_t = \{(\mathbf{x}_{t,j}, y_{t,j})\}_{j=1}^{M_t}$, \\ Number of centroids $K$  % Split long line into multiple lines
            \Statex \textbf{\hspace*{-1.7em} Train Stage: Learning Prototype-based Framework}
            \State Initialize $\mathbf{P} = \{\mathbf{p}^{\text{bona fide}}, \mathbf{p}^{\text{spoof}}\} \in \mathbb{R}^{D \times K \times 2}$
            \For{batch $\mathcal{B}$ in source domains}
              \State $\mathbf{z}_i = f(\mathbf{x}_i)$ for $\mathbf{x}_i \in \mathcal{B}$
              \State Compute losses:
              $\mathcal{L}_{\text{proto}}$, $\mathcal{L}_{\text{con}}^{\text{coarse}}$, $\mathcal{L}_{\text{con}}^{\text{fine}}$, $\mathcal{L}_{\text{orth}}$ 
              \State Update $\mathbf{P}$ and $f$ using combined loss (Eq. 3)
            \EndFor
            \Ensure Learned class prototypes $\mathbf{P} = \{\mathbf{p}^{\text{bona fide}}, \mathbf{p}^{\text{spoof}}\}$
            \Statex \textbf{\hspace*{-1.7em}Test Stage: Source-free Few-shot Adaptation}
            \State Extract features: $\mathbf{Z}_t = \{f(\mathbf{x}_{t,j})\}_{j=1}^{M_t}$
          \end{algorithmic}
        \end{algorithm}
      \end{minipage}
      \hfill
      \begin{minipage}[t]{0.49\textwidth}  % Added [t] for top alignment
        \begin{algorithm}[H]
          \begin{algorithmic}[1]
            \setcounter{ALG@line}{6}
            \Statex \textit{\hspace*{-1.7em}Proposed Method 1: Training-free OT Adaptation}
            \For{$c \in \{\text{bona fide}, \text{spoof}\}$}
              \State Compute $\mathbf{M}_{ij}^c = \|\mathbf{p}_i^c - \mathbf{z}_j^t\|_2^2$; init $\mathbf{a}^c$, $\mathbf{b}^c$  % Split into two lines
              \State $\gamma^{c*} = \argmin_{\gamma \in \Pi(\mathbf{a}^c, \mathbf{b}^c)} \langle \gamma, \mathbf{M}^c \rangle$ $+ \lambda\Omega_{\alpha}(\gamma)$  % Split into two lines
              \State Optimal Transform: $\mathbf{p}^{c*} = T_{\gamma^{c*}}(\mathbf{p}^c)$
            \EndFor
            \Ensure Transformed prototypes $\mathbf{P^*}$ (as classifier)
            \Statex \textit{\hspace*{-1.7em}Proposed Method 2: Lightweight Geodesic mix-up}
            \State Initialize a random linear classifier $l : \mathcal{Z} \rightarrow \mathbb{R}^2$
            \While{not reach target iterations}
            \State $w \sim \text{Beta}(0.4, 0.4)$
            \State $\mu_w = \argmin_{\mu} [w\mathcal{W}(\mu, \mu_s) +$ $(1-w)\mathcal{W}(\mu, \mu_t)]$
            \State Optimize classifier $l$ on $\mathbf{Z}_t \cup \mathbf{P} \cup \{ \mathbf{u}_i \}_{i=1}^{K} \overset{\text{i.i.d.}}{\sim} \mu_w$
            \EndWhile
            \Ensure Classifier $l$ 
          \end{algorithmic}
        \end{algorithm}
      \end{minipage}
      \vspace*{\fill}
    \end{minipage}
  }
\caption{Detailed algorithms for prototype-based backbone training, training-free OT adaptation and light-weight training with Geodesic Mixup.}
\label{algo:main}
\end{figure*}

\section{Additional Visualizations}
Here, we provide more visualizations. As demonstrated in Fig.~\ref{fig:ota_viusal} (left two), OTA (training-free) effectively calibrate prototypes to adapt to target domain while resisting noisy samples. Geodesic Mixup (right two) generates diverse pseudo distributions which respect geometric information of source and target domains.

\section{Scaling Property OTA}
Both training-free and lightweight adaptation methods primarily concern few-shot scenarios where the available target domain data are extremely scarce. In application, it is possible that the few-shot number is moderate. To this end, we scale the few-shot number K from 5 to 50 and test OTA accordingly to evaluate OTA's scaling property. As shown in Table~\ref{tab:scaling}, scaling up few-shot number consistently boost the performance of both training-free and lightweight adaptation OTA methods.

\section{Detailed Algorithms}
Detailed algorithms can be found in Algorithm~\ref{algo:main}.

\clearpage

% \vspace{-0.6cm}  % Tune this value (e.g., -0.5cm to -1cm) to reduce top margin
{
    \small
    \bibliographystyle{ieeenat_fullname}
    \bibliography{main}
}


\end{document}